\documentclass{llncs}
\usepackage{llncsdoc}
\usepackage{amsmath,graphicx}
\usepackage{amssymb}
\usepackage{subfigure}
\usepackage{caption}
\usepackage{tabularx}
\usepackage{graphics}
\usepackage{pifont}
\newcommand{\cmark}{\ding{51}}%
\usepackage{bbm}
\newcommand{\upperRomannumeral}[1]{\uppercase\expandafter{\romannumeral#1}}

\usepackage[]{caption}

\title{Improving Deep Lesion Detection Using 3D Contextual and Spatial Attention}
\author{Qingyi Tao\inst{1,2} \and
Zongyuan Ge\inst{1,3} \and
Jianfei Cai\inst{2} \and
Jianxiong Yin\inst{1} \and
Simon See\inst{1}
}
\authorrunning{Q. Tao et al.}
\institute{NVIDIA AI Technology Center \and
Nanyang Technological University \and
Monash University}

\begin{document}

\maketitle

\begin{abstract}
\label{sec:abstract}

Lesion detection from computed tomography (CT) scans is challenging compared to natural object detection because of two major reasons: small lesion size and small inter-class variation. Firstly, the lesions usually only occupy a small region in the CT image. The feature of such small region may not be able to provide sufficient information due to its limited spatial feature resolution.  Secondly, in CT scans, the lesions are often indistinguishable from the background since the lesion and non-lesion areas may have very similar appearances. To tackle both problems, we need to enrich the feature representation and improve the feature discriminativeness. Therefore, we introduce a dual-attention mechanism to the 3D contextual lesion detection framework, including the cross-slice contextual attention to selectively aggregate the information from different slices through a soft re-sampling process. Moreover, we propose intra-slice spatial attention to focus the feature learning in the most prominent regions. Our method can be easily trained end-to-end without adding heavy overhead on the base detection network. We use DeepLesion dataset and train a universal lesion detector to detect all kinds of lesions such as liver tumors, lung nodules, and so on. The results show that our model can significantly boost the results of the baseline lesion detector (with 3D contextual information) but using much fewer slices. 
\end{abstract}

\section{Introduction}
\label{sec:introduction}

As one of the essential computer-aided detection/diagnosis (CADe/CADx) technologies, lesion detection has been studied by the medical imaging community for automatic disease screening and examination. With the great success of deep convolutional neural network (CNN) adoption in object detection in natural images \cite{ren2015faster,dai2016r}, many researchers utilized CNN based algorithms for detecting diseases in different modalities of medical images such as lesion detection in color retinal images \cite{lam2018retinal}, disease detection in X-ray\cite{li2018thoracic}, etc. 

Despite the progress, lesion detection in computed tomography (CT) images is challenging due to the difficulty of learning discriminative feature representation. The primary reasons accounting for such difficulty include: 1) the lesion size can be extremely small compared to natural objects in a general object detection task, and this degrades the richness in the existing feature representation; 2) the inter-class variance is small, i.e., lesions and non-lesions often have very similar appearances.

To enrich the lesion features, lesion detection methods in CT images often take advantage of the intrinsic 3D context information. The works in \cite{liao2019evaluate,dou2017multilevel} exploit 3D CNNs to encode richer spatial and context information for more discriminative features. However, 3D CNN requires more computational resources and at the same time, it requires more efforts to annotate 3D bounding boxes. This motivates \cite{yan20183d} to use (2+1)D information which aggregates 2D features from multiple consecutive CT slices. 
Particularly, in 3DCE \cite{yan20183d}, a set of consecutive slices
are fed into a 2D detection network to generate feature maps separately, which are then aggregated along the channel dimension for the final detection task. Nevertheless, with the increasing number of neighbouring slices, the information extracted from some of the slices may be irrelevant whereas some of the slices may have higher importance for the correct prediction. 

To solve this problem, we propose an attentive feature aggregation mechanism through a 3D contextual attention module to adaptively select important and relevant slices to be focused on. Moreover, to further improve the feature discrimintiveness for small regions in each CT slice, we introduce a similar spatial attention network as used in \cite{wang2017residual,woo2018cbam} to mine the discriminative regions and concentrate the learning on these regions in each feature map. The self-attentive feature maps could enhance the differentiation between lesion and non-lesion proposals.


\section{Methods}
\label{sec:methods}

\begin{figure}[!t]
\centering
\includegraphics[width=1\linewidth]{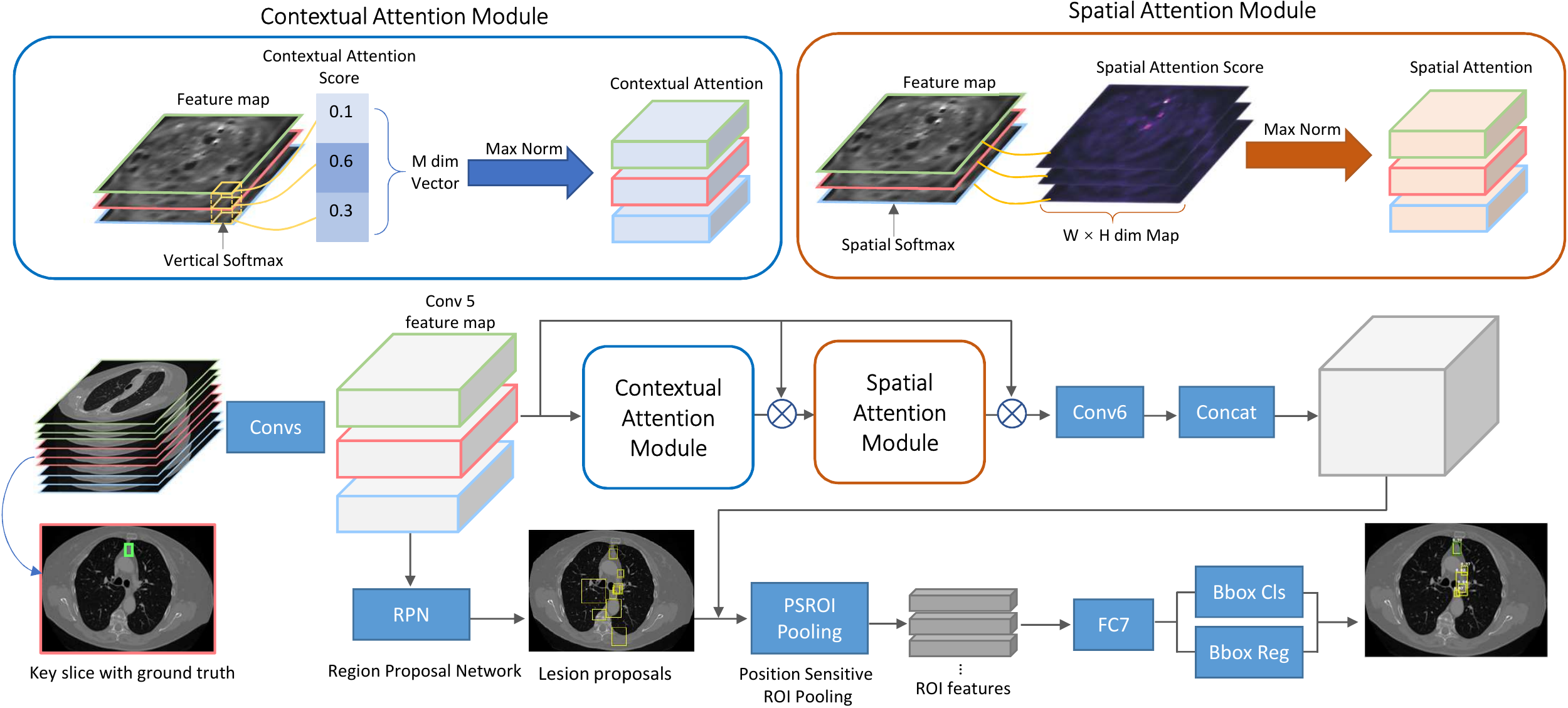}
\caption{Overview of network architecture: Using 3DCE~\cite{yan20183d} as the base framework, we introduce two attention modules: a) contextual attention module to re-weight the feature importance across all input slices; b) spatial attention module to amplify the learning of the most prominent regions within each feature map. }
\label{pic:overview}
\end{figure}

Fig.~\ref{pic:overview} gives an overview of the proposed lesion detection framework. In particular, we adopt an improved version of R-FCN~\cite{dai2016r} as the backbone object detection network which contains a region proposal network (RPN) to generate region proposals, followed by a position-sensitive pooling layer and two fully connected layers. The final classification branch distinguishes between lesion and non-lesion classes and the regression branch refines the bounding box coordinates for more accurate localization. To incorporate 3D context, we group $3M$ slices into $M$ 3-channel images which include the key slice and $(3M-1)$ neighbouring slices before and after the key slice. Similar to~\cite{yan20183d}, instead of performing data-level fusion, we adopt feature integration by forwarding $M$ grouped images to the shared convolutional layers and then concatenating the features right before the position-sensitive ROI (PSROI) pooling layer. Finally, the lesion area is predicted with the aggregated features from all $3M$ slices. The loss functions including the lesion classification loss and the bounding-box regression loss in both RPN and the improved R-FCN are optimized jointly.

The unique part of our framework lies in the introduced two attention modules: contextual attention module and spatial attention module, which are able to extract more important and discriminative features. In the following subsections, we describe our dual-attention mechanism including the contextual attention module to reweight the importance of contextual information, and the spatial attention module to focus on discriminative regions. 


\subsection{3D Contextual Attention Module}

\label{sec:method-ca}
The 3D contextual attention module aims to selectively aggregate features from $M$ input images by attending to the relevant context information among all neighbouring slices. We denote the input feature map ($relu\_5\_3$ for vgg 16) of input image $i$ as $X_i \in \mathbbm{R}^{(D\times W \times H)}$ ($i=[1,2,..M]$),  where $D$ is the number of feature channels, $H$ and $W$ is the height and width of the feature map. The contextual attention module contains a convolutional layer denoted as $\phi_C$ in Equation ~\ref{eq:ca1}. The contextual attention is calculated through a softmax function followed by a normalization operation.
\begin{equation}\label{eq:ca1}
C_i = \phi_{C}(X_i)
\end{equation}
\begin{equation}\label{eq:ca2}
C^{'w,h}_{i,d} =\frac{exp(C^{w,h}_{i,d})}{\sum^{M}_{i=1}exp(C^{w,h}_{i,d})}
\end{equation}
\begin{equation}\label{eq:ca3}
C^{''w,h}_{i,d} = \frac{C^{'w,h}_{i,d}}{\max\limits_{i}|C^{'w,h}_{i,d}|}
\end{equation}
\begin{equation}\label{eq:ca4}
X^{'}_{i} = C^{''}_{i}\otimes{X_{i}}.
\end{equation}

In Equation \ref{eq:ca2}, $C^{w,h}_{i,d}$ is the learnable scalar value which represents the importance or relevance score of each input slice at the position $(w, h)$ in the $d$th feature map of $X_i$. Then a softmax operation is performed along the vertical axis in the dimension of the slice deck (across $M$ slice images) to obtain the contextual attention vector $C^{'w,h}_{d}\in \mathbbm{R}^{M}$. 
This generates attention values between $0$ to $1$ and all the elements sum up to $1$ in $C^{'w,h}_{d}$. Since most elements tend to have small values after the softmax operation, this makes the training sensitive to the learning rate. To stablize the training, as shown in Equation \ref{eq:ca3}, we normalize the attention vector by dividing it with the max element in the vector. Finally, the output feature $X_{i}^{'}$ is obtained by taking element-wise multiplication (denoted as $\otimes$) of the original input features and the attention maps as described in Equation \ref{eq:ca4}. 

With this attention module, the features from different slices are attentively aggregated with a learnable cross-slice attention vector to amplify the relevant contextual features and suppress irrelevant ones.

\subsection{Spatial Attention Module}\label{sec:methodd-sa}

Spatial attention module is designed to optimize the feature learning for prominent regions by applying intra-slice attention on each feature plane. Similar to the contextual attention module, it contains a convolution layer ($\phi_{S}$) followed by a softmax function and a max normalization operation. As shown in Fig.~\ref{pic:overview}, the spatial attention module takes refined features $X_i^{'}$ of all input images and generates spatial attention weight matrix for each feature map. The process can be mathematically written as: 

\begin{equation}\label{eq:sa1}
S_i = \phi_{S}(X_i^{'})
\end{equation}
\begin{equation}\label{eq:sa2}
S^{'w,h}_{i,d} = \frac{exp(S^{w,h}_{i,d})}{\sum^{W}_{w=1}\sum^{H}_{h=1}exp(S^{w,h}_{i,d})}
\end{equation}
\begin{equation}\label{eq:sa3}
S^{''w,h}_{i,d} = \frac{S^{'w,h}_{i,d}}{\max\limits_{w,h}|S^{'w,h}_{i,d}|}
\end{equation}
\begin{equation}\label{eq:sa4}
X^{''}_{i} = S^{''}_{i}\otimes{X^{'}_i}.
\end{equation}

The spatial attention module generates attentive feature maps by amplifying prominent regions within each feature plane in order to improve the richness in features for small lesions and increase the feature discrepancy between lesion and non-lesion regions.

\section{Experiments}

\noindent\textbf{Dataset:} To validate the effectiveness of our approach, we use DeepLesion ~\cite{yan2018deep}  dataset that provides  32,120 axial CT slices with 2D bounding box annotations of lesion regions. The CT images are pre-processed in the same way as that in 3DCE~\cite{yan20183d}. We use the official split of samples which includes $\sim$ 22k samples for training, $\sim$ 5k for validation and another $\sim$ 5k for testing. Following the practice in~\cite{yan20183d}, 35 noisy lesion annotations mentioned in the
dataset are removed for training and testing.

\noindent\textbf{Network and training:} 
We initialize the network using pre-trained ImageNet vgg-16 model. 
In the proposed attention modules, we use a softmax with temperature of 3 for spatial attention and 2 for contextual attention. During the training, each mini-batch has 2 samples and each sample has $M$ three-channel images. Stochastic gradient decent (SGD) with momentum of 0.9 and decay of 5e-5 is used as the optimizer. We train 6 epochs using the base learning rate of 0.001, which is then reduced it by a factor of 10 after the 4th and 5th epochs.

\noindent\textbf{Evaluation metrics:} Following the standard practice, we use intersection-over-union (IoU)$>0.5$ as the measure for overlap to evaluate the prediction results.  We study sensitivities at [0.5, 1, 2, 4, 8, 16] to evaluate the performance by different model variants.

\subsection{Results using Contextual and Spatial Attention }

\begin{table}[!t]

\centering 
\footnotesize
\caption{Sensitivity (\%) at different false false positives (FPs) per image on the test set of the official data
split of DeepLesion. Note that the results of 3DCE are obtained from our experiments which are higher than the reported results in the original paper.}
\begin{tabular*}{\textwidth}{l@{\extracolsep{\fill}}ccccccc}
\hline\hline
Sensitivity @          & 0.5   & 1     & 2     & 4     & 8     & 16    \\ \hline
Improved R-FCN, 3 Slices \cite{yan20183d} &56.5&67.7&76.9&82.8&87.0&89.8\\
3DCE, 9 Slices    \cite{yan20183d}     & 61.7&71.9&79.2&84.3&87.8&89.7 \\
3DCE, 15 Slices  \cite{yan20183d}      &63.0&73.1&80.2&85.2&87.8&89.7 \\
3DCE, 21 Slices  \cite{yan20183d}      &63.2&73.4&80.9&85.6&88.4&90.2\\
\hline

3DCE\_CS\_Att, 9 Slices (Ours) & 67.8&76.3&82.9&86.6&89.3&90.7\\
3DCE\_CS\_Att, 15 Slices (Ours) &70.8&78.6&83.9&87.5&89.9&\textbf{91.4} \\
3DCE\_CS\_Att, 21 Slices (Ours) &\textbf{71.4}&\textbf{78.5}&\textbf{84.0}&\textbf{87.6}&\textbf{90.2}&\textbf{91.4}\\
 \hline
\end{tabular*}\label{tb:sens-full}

\end{table}

\begin{table}[t]
\footnotesize
\caption{ Sensitivity (\%) at 4 FPs per image on the test set of DeepLesion using the baseline model (3DCE) and the proposed model (3DCE\_CS\_Att), both using 15 slices. }
\scriptsize
\begin{tabular*}{\textwidth}{l@{\extracolsep{\fill}}|cccccccc|ccc|cc}
\hline\hline
& \multicolumn{8}{c|}{Lesion type}       & \multicolumn{3}{c|}{Lesion diameter}      & \multicolumn{2}{c}{Slice interval} \\
& LU & ME & LV & ST & PV & AB & KD & BN & \textless{}10 & 10$\sim$30 & \textgreater{}30 & \textless{}2.5    & \textgreater{}2.5   \\ 
\hline
3DCE                    &90.9&88.1&90.4&73.5&82.1&81.3&82.1&\textbf{75.0}&80.9&87.8&82.9&85.8&85.1\\

3DCE\_CS\_Att & \textbf{92.0}&\textbf{88.5}&\textbf{91.4}&\textbf{80.3}&\textbf{85.0}&\textbf{84.4}&\textbf{84.3}&\textbf{75.0}&\textbf{82.3}&\textbf{90.0}&\textbf{85.0}&\textbf{87.6}&\textbf{87.6}\\
\hline
\end{tabular*}\label{tb:sens-split}
\end{table}

Firstly, we evaluate the effectiveness of our overall model, using the framework described in Section \ref{sec:methods}. We compare our method with the baseline methods as shown in Table~\ref{tb:sens-full}. The improved R-FCN uses only 3 input slices fused at data-level. 3DCE improves the performance by adding more neighbouring slices and enabling the feature-level aggregation. Our method improves on 3DCE by introducing the contextual and spatial attention modules for attentive feature aggregation. We reproduce the results of 3DCE with different numbers of slices and achieve slightly higher results than those reported in their paper. Then we evaluate our model using the same numbers of slices as used in 3DCE. The results show that our method constantly boosts the accuracy at various FPs per image by around 7 - 8\% in sensitivity at 0.5 and 2\% in sensitivity at 4.
More surprisingly, it is observed that our model using only 9 slices can greatly outperform the original 3DCE using 21 slices by a large margin while using much less computing resources in terms of GPU memory and computation time.  

We further analyze the detection accuracy on different lesions types and image properties by splitting the test set according to three criteria: 1) Lesion type; 2) Lesion diameter (in mm) and 3) Slice interval (in mm) of the CT scans. The results for each split are shown in Table \ref{tb:sens-split}. There are eight types of lesions, including lung(LU), mediastinum(ME), liver(LV), soft tissue(ST), pelvis(PV), abdomen(AB), kidney(KD), and bone(BN) \cite{yan2018deep}. It is found that our method surpasses 3DCE in almost all lesion types. Especially for soft tissue lesion, our model achieves a large increase of 6.8\% when compared with the baseline. 
In term of lesion diameter, the proposed method is slightly more effective on lesions that are larger than 10 mm. This is probably because the lesions smaller than 10 mm are less than 20 pixels in the CT image and have very low resolution at the attention maps (not greater than $2\times2$ patches). Therefore, the attentive feature enhancement on very small lesions could be less effective than on lesions with slightly larger size.
Lastly, our method can achieve a constant improvement for CT scans with different slice intervals since our model attentively aggregates the relevant information from different slices with a cross-slice normalization.

\subsection{Ablation Study}

\begin{table}[h]

\caption{Sensitivity (\%) at various FPs per image on the test set of the official data split of DeepLesion using different attention components with 15 slices.}
\footnotesize
\begin{tabular*}{\textwidth}{@{\extracolsep{\fill}}cc|ccccccc}
\hline
\hline
C\_Att & S\_Att         & 0.5   & 1     & 2     & 4     & 8     & 16    \\ \hline
                   &                 & 63.0 & 73.1 & 80.2 & 85.2 & 87.8 & 89.7 \\
\cmark     &                 & 64.0 & 74.0 & 81.4 & 86.0 & 88.6 & 90.5 \\
                   & \cmark  & 69.0 & 77.4 & 83.1 & 86.7 & 89.1 & 90.8 \\
\cmark      & \cmark    & \textbf{70.8} & \textbf{78.6} & \textbf{83.9}&\textbf{87.5} & \textbf{89.9} &\textbf{91.4}
\\\hline
\end{tabular*}\label{tb:sens-ablation}
\end{table}

In this section, we investigate the effectiveness of each proposed component. We study the sensitivity at different FPs per image by comparing the baseline 3DCE model with the following variants: 1) adding contextual attention (C\_Att); 2) adding spatial attention (S\_Att); and 3) adding both contextual and spatial attention. As shown in Table \ref{tb:sens-ablation}, applying contextual attention alone brings a slight and constant improvement to the baseline method, whereas the spatial attention module alone performs very well and boosts the sensitivity at 0.5 from 63\% to 69\% with 6\% performance gain. While adding both C\_Att and S\_Att modules, we achieve a higher sensitivity of 70.8\% at 0.5 and further improve the sole S\_Att model by 1.8\%.

It is observed that at fewer FPs per image (0.5, 1), the performance gain is mainly from spatial attention, indicating that the spatial attention is essential to improve the prediction confidence for positive boxes. On the other hand, at higher FPs contextual attention gets more and more important for the performance gain.

\section{Visualization of Detection Results}

\begin{figure}[!t]
\centering
\includegraphics[width=\linewidth]{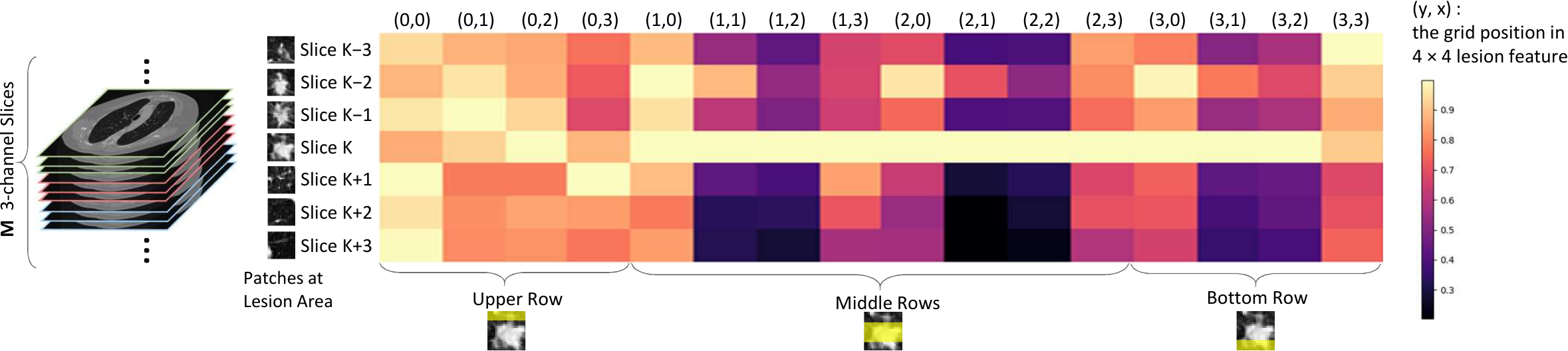}
\caption {Visualization of cross-slice contextual attention vectors based on our model with 7 three-channel images ($M=7$). We visualize the slice patches corresponding to the lesion area in the key slice $K$ as well as its previous and subsequent slices. The lesion region corresponds to $4 \times 4$ grids in Conv 5 feature. Therefore, we obtain 16 attention vectors for each feature grid from the contextual attention module. The vectors are visualized as a heatmap where each column (with a (y, x) coordinate in sub feature map of lesion patch) shows a normalized cross-slice attention vector.}
\label{pic:c_att_vis}
\end{figure}

\begin{figure}[!t]
\centering
\includegraphics[width=1\linewidth]{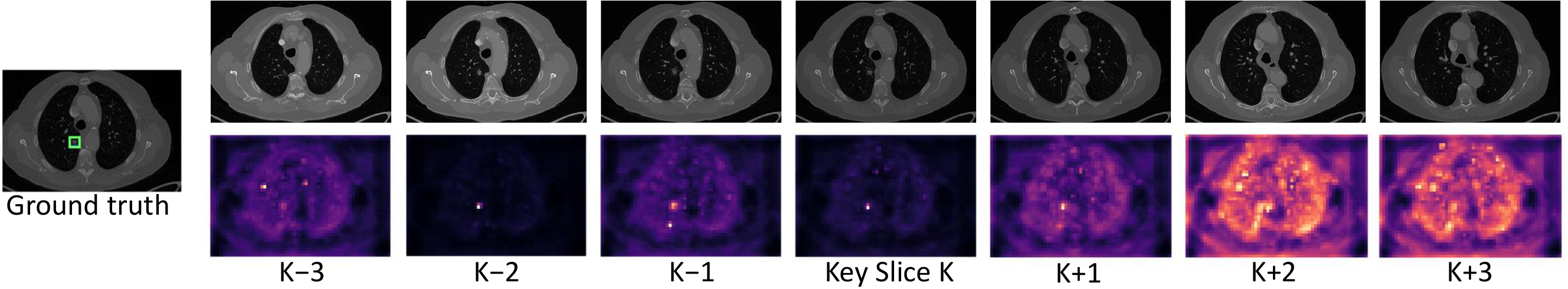}
\caption{ Visualization of spatial attention map based on our model with 7 three-channel images. We can obtain 7 attention maps that are self-normalized to re-weight the feature importance within each feature map. }
\label{pic:s_att_vis}
\end{figure}

\subsection{Qualitative Results and Visualization}

\begin{figure}[t]
\centering
\includegraphics[width=\linewidth]{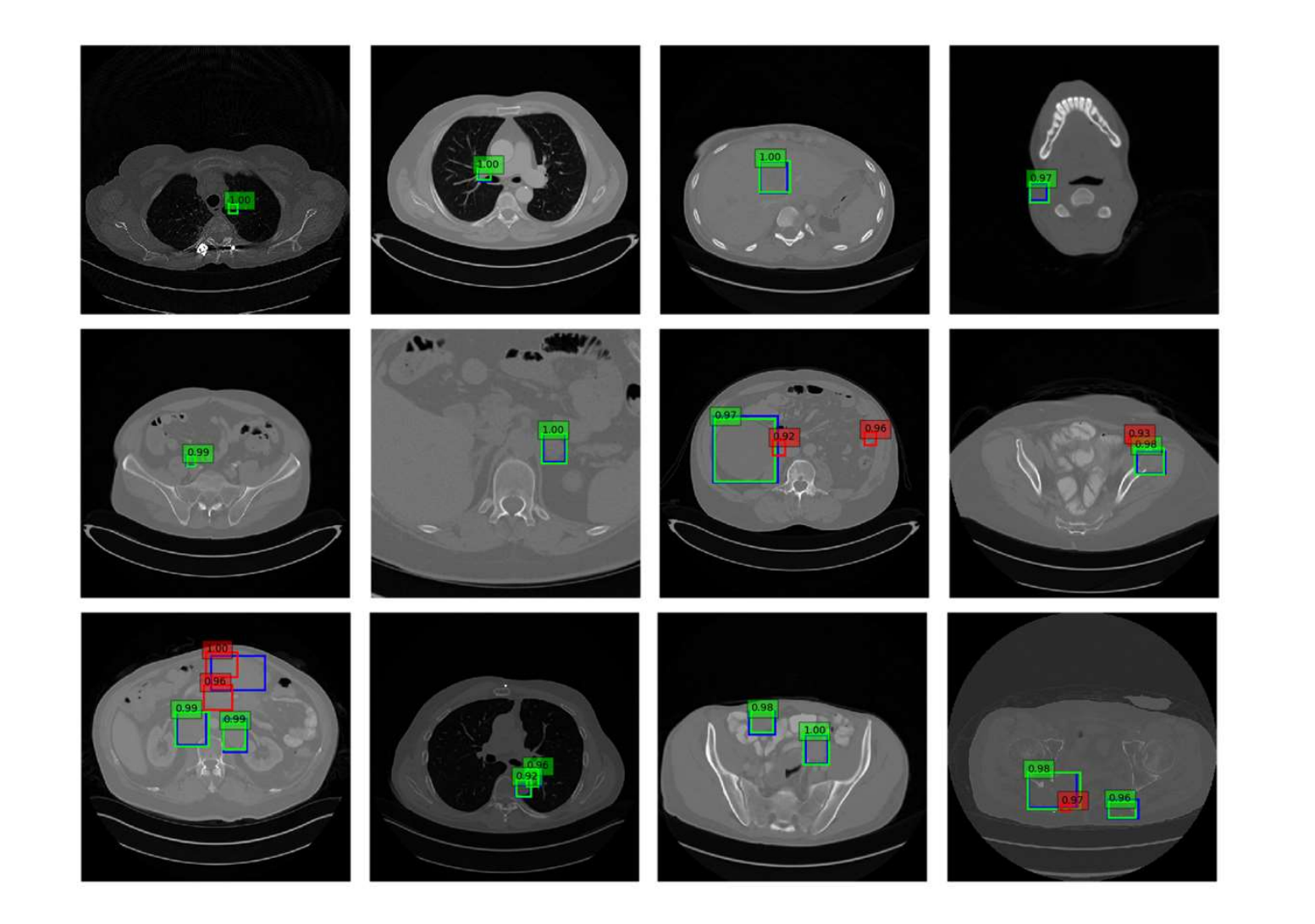}
\caption{Visualization of detection results in official test set of DeepLesion dataset. Blue, green and red boxes represents the ground truth, true positive and false positive boxes respectively.}
\label{pic:res}
\end{figure}

\begin{figure}[t]
\centering
\includegraphics[width=\linewidth]{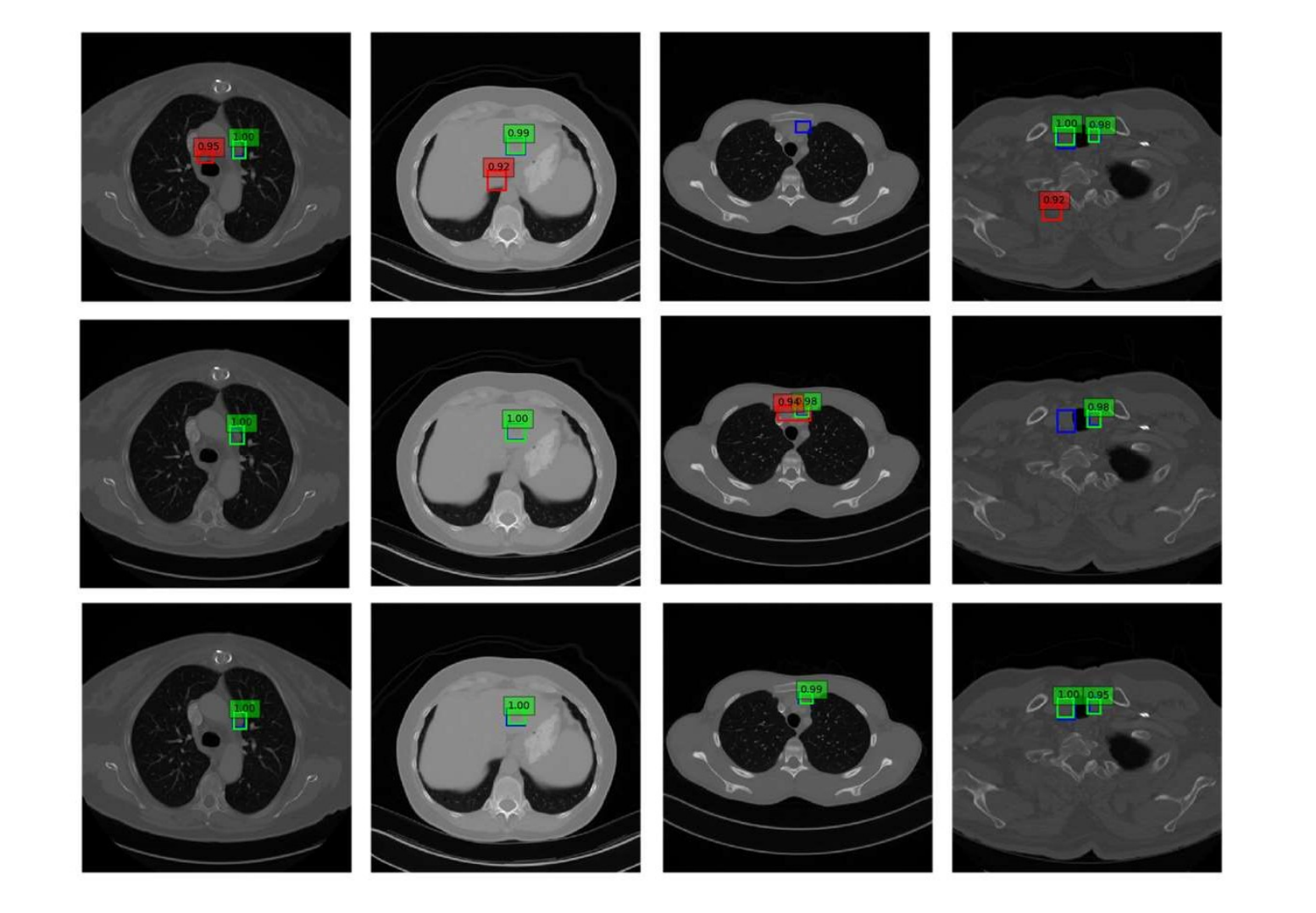}
\caption{Visualization of detection results in official test set of DeepLesion dataset using contextual attention module only (top), spatial attention module only(middle), and both modules (bottom). Blue, green and red boxes represents the ground truth, true positive and false positive boxes respectively.}
\label{pic:res_ablation}
\end{figure}

In this section, to better understand how spatial and contextual attentions work for discriminative region mining, we illustrate some intermediate results of cross-slice contextual attention for soft-sampling and the intra-slice spatial attention for feature re-weighting.
Some examples of detection results are shown in Fig.~\ref{pic:res} and Fig.~\ref{pic:res_ablation}.

\noindent\textbf{Contextual Attention:} We visualize the contextual attention vectors at the lesion area in Fig.~\ref{pic:c_att_vis}. There are $3 \times M$ input slices that are grouped in $M$ 3-channel images. For simplicity, we denote each 3-channel image as a ``Slice'' in the figure. Each column of the heatmap shows a weight vector $v = C^{''y,x}_d \in \mathbbm{R}^M$ to determine the importance of each slice ($K-3$ to $K+3$) at position $(y,x)$ of the lesion feature map. Note that each vector has been normalized by its maximum element.  It shows that at the upper part of the lesion area, all the slices almost have the same importance since there are no prominent features in this area. Near the center of the lesion area, the key slice is given the largest attention. Additionally, Slices $K-1$ and $K-2$ are also well-attended to provide additional information of the lesion since Slices $K-1$ and $K-2$ also catch the lesion appearance. Slices $K+1$, $K+2$ and $K+3$ are suppressed by the contextual attention module since the lesion is absent from these slices.

\noindent\textbf{Spatial Attention:} In Fig.~\ref{pic:s_att_vis}, we show the attention maps generated by the spatial attention module. We use the same example as used in Fig.~\ref{pic:c_att_vis}. Each attention map $S^{''}_{i,d}$ has been normalized by its maximum element. The lesion clearly appears in Slices $K-2$, $K-1$ and key slice $K$. Therefore, in the spatial attention map, we can see a clear pulse at the lesion area. Since the lesion disappears from Slices $K+1$, $K+2$ and $K+3$, the attention maps become plainer by which most feature grids are treated equally.


\section{Conclusion}

In this work, we studied the effectiveness of 3D contextual and spatial attention for lesion detection task in CT scans. The 3D contextual attention has been proposed to attentively aggregate the information from multiple slices. The spatial attention could help to concentrate the feature learning at the most discriminative regions within each feature map.  We validated the effectiveness of our method with various experimental and analytic results, which shows that the proposed method brings a performance boost in the lesion detection task in CT scans.

\bibliographystyle{IEEEbib}
\bibliography{main}

\end{document}